\documentclass[pdflatex, sn-mathphys-num]{sn-jnl}

\usepackage{graphicx}%
\usepackage{multirow}%
\usepackage{amsmath, amssymb, amsfonts}%
\usepackage{amsthm}%
\usepackage{mathrsfs}%
\usepackage[title]{appendix}%
\usepackage{xcolor}%
\usepackage{textcomp}%
\usepackage{manyfoot}%
\usepackage{booktabs}%
\usepackage{algorithm}%
\usepackage{algorithmicx}%
\usepackage{algpseudocode}%
\usepackage{listings}%

\theoremstyle{thmstyleone}%
\theoremstyle{thmstyletwo}%

\theoremstyle{thmstylethree}%

\begin{document}

\title[Article Title]{Balanced Soft mixture-of-expert model for Glaucoma Detection}


\author*[1]{\fnm{Sai Venkatesh} \sur{Chilukoti}}\email{saivenkatesh.chilukoti309@gmail.com}

\author[1]{\fnm{Krishna} \sur{Rauniyar}}\email{krishna.rauniyar1@louisiana.edu}

\author[1]{\fnm{Min} \sur{Shi}}\email{min.shi@louisiana.edu}

\author[1]{\fnm{Xiali} \sur{Hei}}\email{xiali.hei@louisiana.edu}

\affil*[1]{\orgdiv{School of Computing and Informatics}, \orgname{University of Louisiana at Lafayette}, \orgaddress{\street{301 East Lewis Street}, \city{Lafayette}, \postcode{70504}, \state{LA}, \country{USA}}}


\abstract{Glaucoma is a group of eye diseases that damage the optic nerve, often caused by elevated intraocular pressure. It is a leading cause of irreversible vision loss and is typically developed slowly and painlessly, making it difficult to notice until significant damage has occurred. Therefore, early detection is crucial to prevent or slow the progression of vision loss. In recent years, deep learning based uni-modal models have improved the accuracy and efficiency of glaucoma detection, empowering doctors with tools for earlier diagnosis, better monitoring, and timely treatment. Building on this, multi-modal models have emerged, leveraging the strengths of different imaging modalities to learn richer and more robust representations, further enhancing glaucoma detection accuracy. 
However, multi-modal learning faces challenges such as imbalanced and under-optimized uni-modal representations due to joint learning objectives. To address this, we propose a balanced soft mixture-experts model with three experts and load balancing loss. The performance is measured by AUC, our proposed method surpasses the performance of all uni-modal baselines, conventional multi-modal models, and current state-of-the-art balanced multi-modal models. The proposed model can be generalized to other disease detections such as diabetic retinopathy.}




\maketitle

\section{Introduction}\label{sec1}

Glaucoma is the second leading cause of blindness worldwide and affects approximately 4.22 million people in the United States. It can affect people of all ages, but the risk is higher for people over 60 years of age, those with a family history of glaucoma, diabetics, and people with severe nearsightedness~\cite{glaucomaFactsStats2025}. Currently, there is no cure for glaucoma, and vision loss caused by the disease is irreversible. Although medications and surgical treatments can help halt or slow the progression of vision loss, most forms of glaucoma do not exhibit symptoms until the condition has advanced significantly. Therefore, accurate and early detection is crucial to prevent permanent vision loss ~\cite{mayoClinicGlaucoma2025}.  Glaucoma detection generally involves assessing structural damage to the optic nerve along with evaluating visual function. However, clinical examination of the optic nerve head (ONH) and the retinal nerve fiber layer (RNFL) can be subjective and prone to variability. To address this, recent research has focused on developing objective diagnostic tools. Technologies such as confocal scanning laser ophthalmoscopy and optical coherence tomography have been widely studied as complementary methods to support the traditional subjective evaluation of ONH ~\cite{sharma2008diagnostic}. 

Recently, deep learning models for the detection of glaucoma have analyzed images such as fundus photos and OCT scans to automatically detect disease features with high accuracy. They can outperform human experts in some cases and offer scalable solutions for early detection~\cite{phene2019relative}. Most glaucoma detection models employ deep convolutional neural networks (CNNs) specifically designed to analyze particular imaging modalities, such as fundus photographs or OCT scans. These models focus on identifying key indicators of glaucoma, including cup-to-disc ratio, thinning of the retinal nerve fiber layer (RNFL), and morphology of the optic nerve head, which are critical for an accurate diagnosis. To improve performance, advanced techniques such as transfer learning and multiscale feature extraction are commonly used, improving both accuracy and robustness~\cite{christopher2018performance, mahmood2018automated, Li_2020_TMI}. However, despite their strong results, the dependence of uni-modal models on a single type of imaging data can limit their ability to generalize across diverse patient populations and imaging conditions, thus motivating the development of multi-modal approaches that integrate complementary data sources for a more comprehensive detection of glaucoma. 

Recent advances in multi-modal glaucoma detection leverage deep learning models that integrate various imaging modalities to enhance diagnostic accuracy. For instance, Hwang et al. (2025) ~\cite{hwang2025multimodal} developed a neural network using minimally processed fundus photographs, OCT scans, and visual field analyzes, demonstrating superior detection of glaucoma compared to single-mode approaches. Du et al. (2024) ~\cite{du2024hmvgg} introduced the hybrid multi-modulal VGG (HM-VGG) model, which employs attention mechanisms to effectively analyze visual field data, particularly excelling in limited data scenarios. Together, these studies highlight the potential of multi-modal deep learning frameworks to improve early glaucoma detection by combining complementary data sources. Existing multi-modal discriminative models often use a uniform learning objective applied equally to all input modalities. However, this can lead to a problem where some uni-modal representations become under-optimized because one modality dominates the learning process. In other words, the model can focus too much on the most informative modality, causing other modalities to contribute less effectively to overall performance. This imbalance not only reduces the model’s ability to fully leverage complementary information between modalities but also makes it vulnerable to noise or missing data in the dominant modality. Addressing learning imbalance is therefore crucial to develop robust multi-modal models that can integrate diverse data sources effectively and improve generalization. 

In this work, we present a balanced soft mixture-of-experts (SMoE) model designed to address the limitations of existing methods. The architecture comprises three experts, two uni-modal experts, each processing a single modality, and one multi-modal expert that jointly processes both modalities. Expert weights are produced by a CNN-based gating network, which assigns weights to the three experts that sum up to one. To promote equitable contributions from all experts, we incorporate a coefficient-of-variation based load balancing loss. The final prediction is computed as a weighted average of the output of all experts according to the gating network’s assignments.

\section{Results}\label{sec2}

\subsection{Dataset Collection}\label{subsec2} 

\begin{figure}[htbp]
    \centering
    \includegraphics[width=0.8\textwidth]{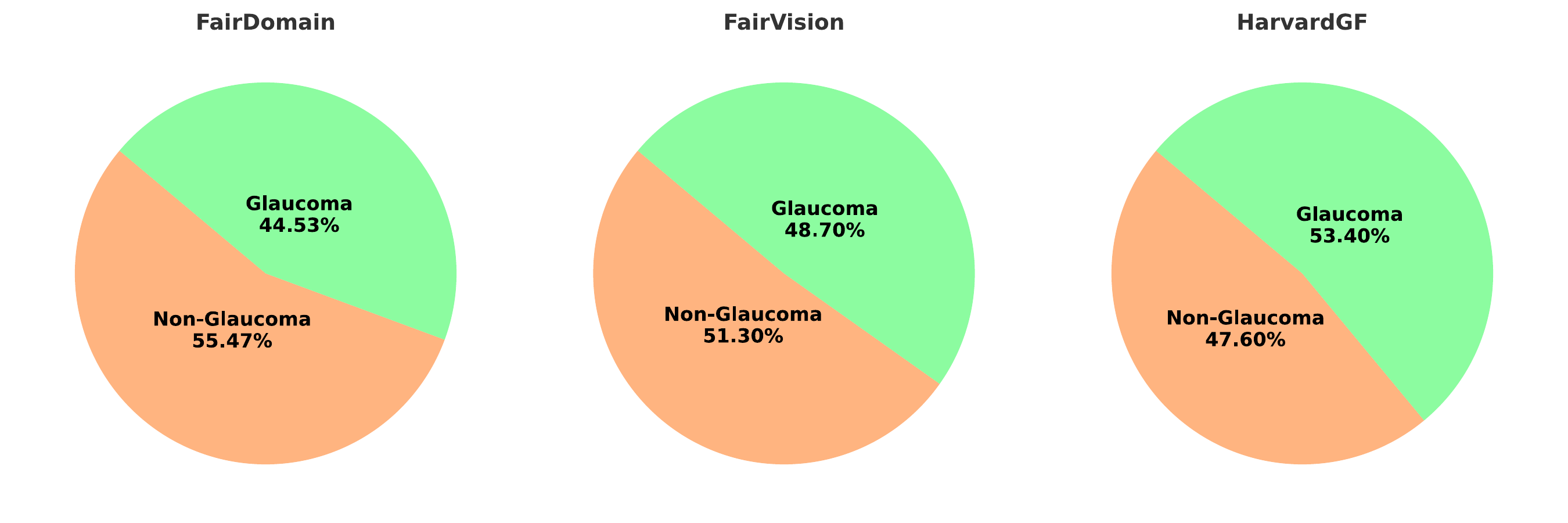}
    \caption{Distribution of glaucoma and non-glaucoma cases across FairDomain, FairVision, and HarvardGF datasets.}
    \label{fig:glaucoma_pie_charts}
\end{figure}

Figure ~\ref{fig:glaucoma_pie_charts} illustrates the distribution of glaucoma and non-glaucoma cases across the FairDomain, FairVision, and HarvardGF datasets. Each pie chart represents the proportion of subjects labeled with glaucoma versus those without it, based on visual field test results. In FairDomain, 44.53\% of the subjects were diagnosed with glaucoma, while 55.47\% were not. The FairVision dataset shows a slightly higher prevalence of glaucoma at 48.7\%, with non-glaucoma subjects comprising 51.3\% of the population. In contrast, HarvardGF contains the highest proportion of glaucoma cases at 53.4\%, with non-glaucoma subjects making up 47.6\%. These distributions highlight the variability in disease prevalence across datasets, which is important to consider when evaluating model performance and generalizability in glaucoma detection tasks.

In this study, we used three datasets FairVision, FairDomain, and HarvardGF, each comprising paired samples per subject to support multi-modal learning. The FairVision dataset~\cite{Luo2023FairVision} includes 30,000 subjects, with 10,000 each for age-related macular degeneration (AMD), diabetic retinopathy (DR), and glaucoma. Among these, 48.7\% are diagnosed with glaucoma and 51.3\% are considered normal. All subjects underwent visual field (VF) testing, which can bias the sample toward more glaucoma cases; however, VF-based labeling is more reliable and consistent than clinician judgment. The demographic characteristics of the FairVision dataset are as follows: the average age is 64.1 ± 17.0 years; gender distribution is 57.1\% female and 42.9\% male; racial composition includes 78.6\% White, 13.7\% Black, and 7.7\% Asian; 3.8\% of subjects identify as Hispanic and 96.2\% as non-Hispanic; preferred language is 91.6\% English, 1.8\% Spanish, and 5.9\% other languages; and marital status includes 24.6\% single, 7.0\% divorced, 0.9\% legally separated, and 8.6\% widowed. Each dataset includes three types of data: (1) Retinal imaging data, which consists of both 2D scanning laser ophthalmoscopy (SLO) fundus images that capture the retinal surface and 3D optical coherence tomography (OCT) scans that measure the in-depth structure of retinal layers. Both imaging modalities are effective for assessing retinal abnormalities caused by eye diseases, although OCT is generally regarded superior in diagnostic accuracy; (2) demographic identity group information; and (3) disease diagnoses for the three major retinal conditions AMD, DR, and glaucoma that together affect more than 380 million people worldwide.  

The FairDomain~\cite{Tian2024FairDomain} dataset includes subjects tested between 2010 and 2021 at a large academic eye hospital affiliated with Harvard Medical School. It contains data for both medical segmentation and classification tasks. For medical classification, the dataset comprises four types of data: (1) en-face fundus imaging scans, (2) scanning laser ophthalmoscopy (SLO) fundus imaging scans, (3) patient demographic information, and (4) glaucoma diagnoses. Subjects in this dataset are classified into two categories—normal and glaucoma—based on visual field test results. The classification subset consists of 10,000 samples from 10,000 individual subjects, with an average age of 60.9 ± 16.1 years. This dataset is split into 8,000 samples for training and 2,000 for testing. It includes six demographic attributes: age, gender, race, ethnicity, preferred language, and marital status. The gender distribution is 72.5\% female and 27.5\% male. The racial distribution includes 76.8\% White, 14.5\% Black, and 8.7\% Asian. The ethnicity is reported as 96.0\% non-Hispanic and 4.0\% Hispanic. The preferred language includes 92.6\% English, 1.7\% Spanish, 3.6\% other languages, and 2.1\% unknown. The marital status distribution is as follows: 58.5\% married or partnered, 26.1\% single, 6.9\% divorced, 0.8\% legally separated, 1.9\% widowed, and 5.8\% unknown. 

The HarvardGF~\cite{Luo2024HarvardGF} dataset comprises subjects tested between 2010 and 2021 in a large academic eye hospital. It includes three types of data: (1) optical coherence tomography (OCT) scans, (2) patient demographic information, and (3) glaucoma diagnosis based on visual field (VF) tests. The OCT data consist of 2D retinal nerve fiber layer thickness (RNFLT) maps and 3D OCT-B scans. OCT is a state-of-the-art 3D imaging technique widely used to diagnose eye diseases such as glaucoma, age-related macular degeneration (AMD), and diabetic retinopathy (DR). For glaucoma assessment, the 2D RNFLT map is derived from the 3D OCT-B scans and represents the vertical distance between the inner limiting membrane and the retinal nerve fiber layer boundaries, segmented using the manufacturer’s software. This 2D RNFLT map, a standard structural measurement in clinical practice, has a resolution of $200 \times 200$ pixels and covers a $6 \times 6~\text{mm}^2$ area centered on the optic disc, with RNFLT values ranging from 0 to 350 microns. To ensure high image quality, OCT scans with a signal strength below 6 (on a scale where 10 denotes the highest quality) were excluded. The dataset also includes 3D OCT-B scans to complement the 2D structural data. Demographic data encompass age, gender, race, ethnicity, language proficiency, and marital status. As the dataset is specifically curated to study racial fairness, the number of subjects was balanced across racial groups using the hospital’s dataset. Glaucoma diagnoses were determined based on the VF test performed using the Humphrey Field Analyzer, which assesses visual sensitivity within a 24-degree radius from the central fixation point in each eye. Only reliable VF tests, as defined by clinically accepted quality control criteria and manufacturer guidelines, were included to maintain diagnostic consistency. 

\subsection{Glaucoma Detection Results} 

\begin{table}[h]
\caption{Overall AUC ($\uparrow$) for three datasets, grouped by method and modality for Uni-Modal}
\label{tab:grouped_auc}
\begin{tabular*}{\textwidth}{@{\extracolsep\fill}lll c}
\toprule
Dataset & Method & Modality Type & Overall AUC \\
\midrule

\textbf{Fairvision} 
& \multirow{2}{*}{DenseNet}       & Slo Fundus      & 78.94$\pm$0.0080 \\
&                                 & Oct Bscans      & 80.86$\pm$0.0133 \\
& \multirow{2}{*}{EfficientNet}   & Slo Fundus      & 78.74$\pm$0.0055 \\
&                                 & Oct Bscans      & 81.16$\pm$0.0056 \\
& \multirow{2}{*}{ResNet}         & Slo Fundus      & 76.32$\pm$0.0043 \\
&                                 & Oct Bscans      & 81.90$\pm$0.0046 \\
& \multirow{2}{*}{ViT-B}          & Slo Fundus      & 79.52$\pm$0.0033 \\
&                                 & Oct Bscans      & 81.78$\pm$0.0046 \\
\midrule

\textbf{Fairdomain} 
& \multirow{2}{*}{DenseNet}       & Slo Fundus      & 80.84$\pm$0.0064 \\
&                                 & Oct Fundus      & 81.70$\pm$0.0045 \\
& \multirow{2}{*}{EfficientNet}   & Slo Fundus      & 80.02$\pm$0.0058 \\
&                                 & Oct Fundus      & 81.50$\pm$0.0050 \\
& \multirow{2}{*}{ResNet}         & Slo Fundus      & 79.36$\pm$0.0015 \\
&                                 & Oct Fundus      & 79.24$\pm$0.0120 \\
& \multirow{2}{*}{ViT-B}          & Slo Fundus      & 71.86$\pm$0.0048 \\
&                                 & Oct Fundus      & 69.24$\pm$0.0038 \\
\midrule

\textbf{HarvardGf} 
& \multirow{2}{*}{DenseNet}       & Rnflt           & 84.94$\pm$0.0062 \\
&                                 & Oct Bscans      & 78.52$\pm$0.0259 \\
& \multirow{2}{*}{EfficientNet}   & Rnflt           & 85.32$\pm$0.0018 \\
&                                 & Oct Bscans      & 78.60$\pm$0.0177 \\
& \multirow{2}{*}{ResNet}         & Rnflt           & 83.66$\pm$0.0127 \\
&                                 & Oct Bscans      & 79.78$\pm$0.0081 \\
& \multirow{2}{*}{ViT-B}          & Rnflt           & 83.42$\pm$0.0055 \\
&                                 & Oct Bscans      & 79.94$\pm$0.0096 \\
\bottomrule
\end{tabular*}
\end{table}

Table~\ref{tab:grouped_auc} presents the overall Area Under the Curve (AUC) scores achieved by the baseline models in the three datasets. FairVision, HarvardGF, and FairDomain. The baseline models evaluated include DenseNet~\cite{Huang2017DenseNet}, EfficientNet~\cite{Tan2019EfficientNet}, ResNet~\cite{He2016ResNet}, and ViT-B~\cite{Dosovitskiy2021ViT}. Each model was trained and tested on individual imaging modalities within each dataset to assess performance consistency and generalizability. Across all datasets and modalities, EfficientNet consistently demonstrates strong performance, often achieving the highest or near-highest AUC values. In the FairVision dataset, EfficientNet yields the highest AUC of 81.16\% on OCT-B scans, outperforming other models in this modality. Similarly, in FairDomain, EfficientNet performs competitively, particularly on OCT fundus images with an AUC of 81.50\%, closely following DenseNet. In particular, on the HarvardGF dataset, EfficientNet achieves the best overall performance with an AUC of 85.32\% and 78.60\% on RNFLT maps and OCT-B scans, indicating its robustness in handling structural retinal measurements. The superior performance of EfficientNet can be attributed to its compound scaling method, which systematically and uniformly scales the depth, width, and input resolution. This design principle allows EfficientNet to balance the trade-off between model complexity and representational capacity, leading to enhanced generalization across diverse imaging modalities and datasets.

In contrast, ResNet, while a widely used and proven architecture, demonstrates more variable performance. In the FairVision dataset, ResNet achieves a relatively competitive AUC of 81.90\% on OCT-B scans, slightly outperforming EfficientNet in that specific modality. However, its performance deteriorates in FairDomain, particularly in OCT fundus images where the AUC drops to 79.24\%, indicating potential limitations in capturing modality-specific retinal characteristics. Furthermore, although ResNet performs reasonably well in the HarvardGF dataset, achieving 83.66\% on RNFLT and 79.78\% on OCT B-scans, it does not match the consistency observed with EfficientNet. ViT-B, while representing a modern transformer based approach, shows the most variability among the models, especially in FairDomain, where it records a notable decline in AUC (69.24\%) for OCT fundus images. This suggests a higher sensitivity to dataset characteristics and modality, possibly due to its reliance on larger data volumes and positional encoding mechanisms, which may not generalize as effectively in limited medical datasets. Overall, these results indicate that EfficientNet is the most robust and reliable model for uni-modal retinal image analysis. Its consistently high performance across modalities and datasets underscores its ability to extract discriminative features while maintaining computational efficiency.

\begin{table}[h]
\caption{Overall AUC ($\uparrow$) for three datasets, grouped by method and modality for Multi-Modal}
\label{tab:grouped_auc_MM}
\begin{tabular*}{\textwidth}{@{\extracolsep\fill}lll c}
\toprule
Dataset & Method & Overall AUC \\
\midrule

\textbf{Fairvision} 
& Normal          & 81.86$\pm$0.0113 \\
& OPM         & 82.62$\pm$0.0057 \\
& OGM      & 82.18$\pm$0.0078 \\
& CGGM   & 81.98$\pm$0.0092 \\
& OPM+OGM          & 82.54$\pm$0.0035 \\
& \textbf{SMoE}          & \textbf{83.29$\pm$0.0045} \\
\midrule

\textbf{Fairdomain} 
& Normal          & 83.58$\pm$0.0031 \\
& OPM         & 84.28$\pm$0.0027 \\
& OGM       & 83.94$\pm$0.0086 \\
& CGGM   & 83.22$\pm$0.0028 \\
& OPM+OGM            & 84.30$\pm$0.0048 \\
& \textbf{SMoE}          & \textbf{84.31$\pm$0.0029}\\
\midrule

\textbf{HarvardGf} 
& Normal          & 85.60$\pm$0.0056 \\
& OPM       & 85.30$\pm$0.0048 \\
& OGM       & 85.48$\pm$0.0045 \\
& CGGM   &  84.96$\pm$0.0088 \\
& OPM+OGM            & 85.46$\pm$0.0045 \\
& \textbf{SMoE}          & \textbf{86.33$\pm$0.0026} \\
\bottomrule
\end{tabular*}
\end{table}

Table~\ref{tab:grouped_auc_MM} presents the overall AUC results for various multi-modal (MM) learning approaches across the FairVision, FairDomain, and HarvardGF datasets. The "Normal" baseline refers to a conventional multi-modal model trained without any explicit strategy for balancing modality contributions. OPM (On-the-fly Prediction Modulation)~\cite{wei2024onthefly} represents a forward-pass modulation technique designed to promote balanced learning between modalities. In contrast, OGM (On-the-fly Gradient Modulation)~\cite{Peng2022BalancedOGM} adjusts the gradient flow during backpropagation to address the imbalance in learning dynamics. CGGM (Classifier-Guided Gradient Modulation)~\cite{Guo2024CGGM} extends this idea by using class-specific gradient signals to guide balanced optimization across modalities. The combined method, OPM $+$ OGM, integrates prediction modulation in the forward pass with gradient modulation in the backward pass to enhance learning balance throughout the training process. Inspiring from the DynMM~\cite{Xue2023DynMM}, we propose the SMoE (Soft Mixture-of-Experts), which employs a soft expert assignment mechanism with a load-balancing objective to encourage uniform expert utilization across different data modalities. This method aims to adaptively specialize expert branches while preventing mode collapse or expert under utilization, resulting in improved and more robust multi-modal learning. 

Overall, most of the balanced multi-modal strategies demonstrate improvements over the conventional baseline model. However, there are notable exceptions where certain methods fail to outperform the baseline. Specifically, CGGM performs less poorly than the normal model in FairDomain (83.22\% vs. 83.58\%) and HarvardGF (84.96\% vs. 85.60\%), while OPM performs slightly worse in HarvardGF (85.30\% vs. 85.60\%). Furthermore, OGM shows a performance drop in FairDomain (83.94\%) compared to the normal model (83.58\%). These results suggest that, while balancing mechanisms can enhance learning dynamics, their effectiveness may be dataset dependent and not generalizable in all settings. Among modulation based approaches, the combined method OPM $+$ OGM achieves the most consistent improvements, slightly outperforming individual components in FairVision (82.54\%) and FairDomain (84.30\%), and remains competitive on HarvardGF (85.46\%). This indicates that jointly optimizing forward and backward learning dynamics contributes to more stable and generalizable performance across datasets. The best performing method in all datasets is the proposed SMoE model, which consistently achieves the highest AUC scores: 83.29\% in FairVision, 84.31\% in FairDomain, and 86.33\% in HarvardGF. These results reflect not only strong predictive performance, but also a high degree of robustness across different data distributions and modality combinations. The consistent gains observed with SMoE suggest that its design effectively mitigates modality imbalance while leveraging diverse expert representations, leading to a more balanced and adaptive multi-modal learning process.   

\section{Discussions} \label{sec3} 

Deep learning approaches have been increasingly adopted to analyze clinical data such as VF, fundus image, and OCT scans for automated glaucoma detection~\cite{Russakoff2020GlaucomaOCT, Christopher2020DeepLearningGlaucoma}. Recent developments in multi-modal learning have advanced glaucoma detection by integrating information from diverse modalities such as fundus photography, optical coherence tomography (OCT), visual field data, and clinical metadata. These approaches exploit the complementary nature of structural and functional data to improve diagnostic accuracy. Hwang et al.~\cite{hwang2025multimodal} proposed a neural network that jointly learns from fundus images, OCT, and visual field tests using modality-specific encoders and a fusion network to improve disease classification. Huang et al. ~\cite{Huang2022FundusVFfusion} introduced a probabilistic deep learning framework that combines fundus and visual field modalities to improve the reliability of the model. Li and Pun~\cite{Li2023ELF} designed ELF, a fusion model that integrates local and global features of OCT and fundus modalities. Cai et al. ~\cite{Cai2022COROLLA} proposed COROLLA, which uses contrastive learning to align multi-modal representations from fundus and OCT-derived maps. Li et al. ~\cite{Li2023GMNNnet} further extended multi-modal modeling by incorporating fundus images, segmentation features, and patient metadata through a three-branch architecture. These methods collectively highlight the value of multi-modal learning in capturing complex pathological signatures, enhancing generalization, and improving clinical relevance in glaucoma detection systems.  

In multi-modal glaucoma detection, imbalanced learning occurs when certain modalities, typically those with stronger or more easily optimized features, dominate the training process, while others contribute minimally. This imbalance leads to the under utilization of valuable complementary information and limits the model’s ability to fully capture the disease’s complexity. The issue arises during both the forward and backward phases of training: dominant modalities produce overconfident predictions and receive disproportionately large gradient updates, while weaker modalities are overshadowed. To address this, balanced multi-modal learning strategies introduce mechanisms such as on-the-fly prediction modulation, which suppresses overconfident predictions from dominant modalities, and gradient modulation, which adjusts the learning signal to ensure that all modalities are fairly trained. CGGM addresses this issue by introducing a classifier-aware mechanism that modulates gradients during backpropagation, ensuring that each modality contributes proportionately to the learning process. By aligning gradients based on both their strength and directional agreement with the classifier’s objective, CGGM promotes stable optimization and encourages more uniform feature learning. These strategies promote more equitable representation learning, leading to improved generalization, robustness, and clinical reliability in diverse patient cases. 

The proposed soft mixture-of-experts (SMoE) model with three experts offers several key advantages over conventional balanced multi-modal learning strategies. First, it enables modality-specific specialization, allowing each expert to focus on learning representations from distinct input types such as SLO, OCT, or their combination. This design contrasts with shared representation models, which often struggle to capture modality-specific nuances. Second, SMoE employs a soft expert routing mechanism that adaptively determines the contribution of each expert based on the input, allowing the model to dynamically adjust its reliance on different modalities at a per-sample level. This flexibility is especially beneficial for handling the inherent variability in clinical imaging data. Lastly, SMoE demonstrates robustness to modality noise or dropout, as it does not assume equal importance of all modalities for every instance. Instead, it can reduce noisy or less informative modalities, making it particularly suitable for real-world settings where data quality and completeness can vary significantly across patients. The performance of SMoE, as shown in Table~\ref{tab:grouped_auc_MM}, reflects these architectural advantages. SMoE consistently achieves the highest overall AUC in the three datasets, with 83.29\% in FairVision, 84.31\% in FairDomain, and 86.33\% in HarvardGF. These results surpass those of all baseline and balanced multi-modal learning methods, including OPM, OGM, CGGM, and their combinations. Consistent gains in datasets underscore the model’s ability to generalize well despite differences in data distribution and modality characteristics. In additon, the narrow standard deviations observed indicate stable performance between runs, reinforcing SMoE’s robustness. Together, these findings validate the effectiveness of soft expert specialization and adaptive routing in improving both accuracy and reliability in multi-modal clinical prediction tasks. 

However, the proposed equitable model and its evaluation come with several limitations. First, the model does not explicitly optimize for demographic fairness, which may result in inconsistent fairness outcomes across different demographic subgroups. Second, the SMoE design in this study is based primarily on the EfficientNet backbone, chosen due to its superior performance compared to other baseline models. However, it would be valuable to explore more powerful model architectures and training paradigms, such as adapting pre-trained foundation models~\cite{Nguyen2023LVMMed, Gupta2025MedMAE, Sowrirajan2021MoCoCXR, Zhou2023RETFound}, which could improve performance through low-rank adaptation techniques~\cite{Zanella2024LowRankFSL}. Third, SMoE introduces additional model complexity by incorporating multiple expert branches and a gating mechanism, leading to a larger parameter space and increased computational requirements compared to conventional balanced multi-modal approaches~\cite{Shazeer2017SparselyGatedMoE, Yan2023MoESurvey}. This added complexity may hinder deployment in resource-constrained clinical environments. Finally, SMoE assumes that all data modalities are available during training, which may not be the case in real-world settings. When modality data are missing, additional mechanisms such as modality imputation or masked expert dropout would be needed, which could increase training complexity and raise concerns about model robustness ~\cite{Baltrušaitis2019MultimodalSurvey}. 

In conclusion, this work presents a comprehensive framework for addressing modality imbalance in multi-modal glaucoma detection through a soft mixture-of-experts (SMoE) model. By enabling modality-specific specialization and adaptive expert routing, SMoE effectively mitigates the limitations of prior balanced learning strategies and consistently outperforms baseline methods across multiple datasets. The results highlight its robustness, generalizability, and clinical relevance in handling diverse and variable imaging data. While the model introduces additional complexity and relies on complete modality availability, its design offers a promising direction for equitable and high-performing multi-modal learning. Future work will focus on improving model efficiency, incorporating explicit fairness constraints, and adapting to foundation model-based paradigms for broader clinical applicability.

\section{Methods} \label{sec4}

\subsection{Comparitive methods}

In this study, we have used four balanced multi-modal methods, OPM, OGM, CGGM, and the combination of OPM and OGM. In the following, we briefly describe the OPM, OGM, and CGGM methods.  All of these methods use the EfficientNet B1 model as the backbone.

\begin{figure}[ht]  
  \centering
  \includegraphics[width=0.8\textwidth]{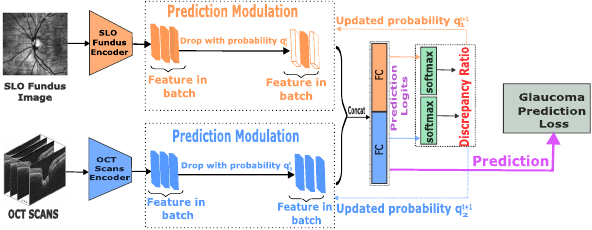}
  \caption{The pipeline of the On-the-fly Prediction Modulation. Here we take SLO Fundus and OCT SCANS. In the feed-forward stage, the feature of modality m (SLO Fundus) is randomly dropped with probability $q^{m}$, where the probability is determined by the discriminative discrepancy ratio at the last iteration. Via OPM, the remained feature of suppressed modality could affect the multi-modal prediction more, accordingly improving its learning.} 
  \label{fig2}
\end{figure}

On-the-fly Prediction Modulation (OPM) is a method designed to balance the contributions of different modalities during multi-modal learning. During training, it evaluates how dominant each modality is by measuring its discriminative power (i.e., how well it can predict the target on its own). If a modality is found to be too dominant, OPM randomly drops its features in the forward pass with a probability proportional to its dominance. This encourages the model to pay more attention to the weaker modalities, leading to more balanced feature learning across the modalities and better overall performance. Figure~\ref{fig2} describes the OPM pipeline. 

\begin{figure}[ht]
  \centering
  \includegraphics[width=0.8\textwidth]{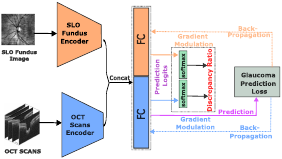}
  \caption{The pipeline of the On-the-fly Gradient Modulation strategy. Here we take SLO Fundus and OCT SCANS. In the back-propagation stage, the gradient of modality m is modulated with km, which is determined by the discriminative discrepancy ratio at this iteration. Via OGM, the gradient of modality with more discriminative information is weakened, while the remained modality is not affected and can gain more training.}
  \label{fig3}
\end{figure}

On-the-fly gradient modulation with Gaussian enhancement (OGM\_GE) combines two techniques to address modality imbalance during multi-modal training. First, On‑the‑fly Gradient Modulation (OGM) dynamically reduces the optimization emphasis on dominant modalities by scaling down their gradients based on their uni-modal performance discrepancy. Second, Gaussian enhancement (GE) adds adaptive Gaussian noise to modulated gradients to restore stochasticity lost by suppression and improve generalization. Together, OGM\_GE ensures that weaker modalities become more optimized while preserving overall robustness and boosting both uni-modal and multi-modal performance. Figure~\ref{fig3} describes the OGM pipeline. 

\begin{figure}[ht]
  \centering
  \includegraphics[width=0.8\textwidth]{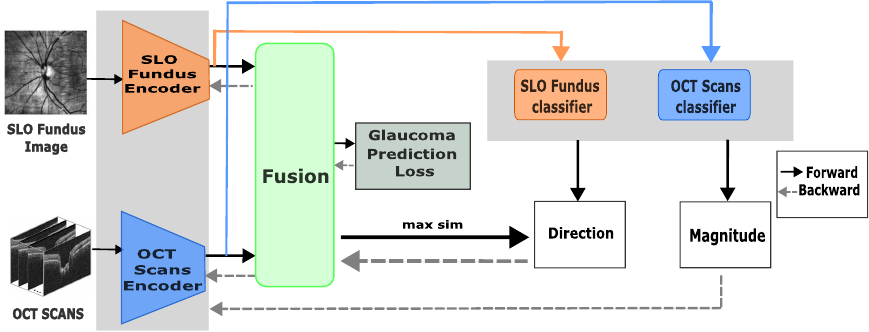}
  \caption{The overall architecture of CGGM. During the training stage, classifiers are introduced to calculate the directions of uni-modal gradients. During the inference stage, the classifiers are discarded.}
  \label{fig4}
\end{figure}

Classifier-Guided Gradient Modulation (CGGM) is a technique proposed by Guo \textit{et al.}. (2024) for balancing multi-modal training by adaptively adjusting each modality’s contribution in backpropagation. Unlike prior methods that only scale the magnitude of the gradient, CGGM also considers the direction of the gradient, using guidance from a classifier to align or alter modality-specific gradients so that they contribute constructively toward the final objective of the task. It thereby curbs over-reliance on dominant modalities and encourages underutilized ones to influence learning more effectively. Figure~\ref{fig4} describes the architecture of CGGM.  

\subsection{The Proposed Balanced Soft Mixture of expert model}

The proposed approach consists of three experts, namely the SLO Fundus expert, OCT Scans expert, and multi-modal expert (considering the modalities from Fairvision dataset). Specifically, all of these experts use the EfficientB1 neural network as the backbone.EfficientNet-B1 is a member of the EfficientNet family, which is known to achieve state-of-the-art accuracy while maintaining exceptional computational efficiency. The architecture builds on a baseline network, EfficientNet-B0, discovered through neural architecture search~\cite{Tan2019MnasNet}, and scales it in a balanced and principled manner to enhance performance. EfficientNet-B1 represents the first scaled-up variant in the series, offering improvements in depth, width, and resolution without disproportionately increasing computational cost. The architecture of EfficientNet-B1 is organized into seven distinct blocks, each designed to progressively extract higher-level visual features. These blocks consist of multiple layers of Mobile Inverted Bottleneck Convolution (MBConv), originally introduced in MobileNetV2~\cite{Sandler2018MobileNetV2}. Each MBConv layer follows an inverted structure in which the input is first expanded to a higher-dimensional space, processed using efficient depthwise separable convolutions~\cite{Howard2017MobileNets}, and then projected back to a lower dimension. To further boost the representational power, each MBConv includes squeeze-and-excitation modules~\cite{Hu2018SENet}, which dynamically recalibrate channel-wise feature responses and help the network focus on the most informative elements of the input. EfficientNet-B1 begins with a standard convolution and max-pooling layer, which prepares the input for deeper processing. As input progresses through the blocks, the network gradually increases the number of filters and kernel sizes, allowing it to transition from capturing fine-grained details in the early layers to more abstract semantic patterns in the later stages. This hierarchical flow of information supports efficient and robust feature extraction in varying image complexities. The defining feature of EfficientNet is its compound scaling method, which uniformly scales the network depth (number of layers), width (number of channels), and input resolution in a coordinated way. Unlike traditional approaches that scale only one dimension, compound scaling adjusts all three in a balanced fashion, ensuring that the model's increased capacity is utilized effectively. EfficientNet-B1 applies this method with a moderate scaling factor, making it larger and more powerful than EfficientNet-B0 while still maintaining practical runtime performance. This strategy allows the network to preserve its efficiency gains while delivering improved accuracy across a range of image classification tasks.  

Although EfficientNet excels at extracting hierarchical features from a single-modality input, it is inherently limited to the information content of that modality alone. In contrast, multi-modal learning integrates complementary information from diverse data sources, such as combining fundus images with OCT scans in ophthalmology, allowing the model to capture a richer and more discriminative representation of the underlying features. This fusion enhances robustness, improves generalization across varying conditions, and often leads to superior performance in complex visual recognition tasks where single-modality cues may be insufficient. Balancing contributions from different modalities is a key challenge in multi-modal learning. Methods like On-the-fly Prediction Modulation (OPM) and On-the-fly Gradient Modulation (OGM) attempt to address this by modulating forward activations or gradients based on modality discrepancy power. However, OPM improves inference but ignores gradient imbalance, while OGM corrects gradients but may suppress strong modalities. Classifier-Guided Gradient Modulation (CGGM) adjusts the gradient flow to each modality based on the softmax confidence of its corresponding classifier, promoting stronger learning from more confident modalities. However, when classifiers produce similar or low confidence scores, especially under ambiguous inputs, CGGM can introduce noise or instability in gradient updates, reducing its reliability in balancing learning across modalities. Even OPM+OGM shows inconsistent behavior across tasks. To address these issues, we introduce a soft Mixture-of-Experts (soft MoE) model that assigns adaptive, learnable weights to modality-specific experts. 

The proposed balanced soft MoE is shown in figure [5]. It contains the gating network and three experts. First, the input (two image modalities) is given to the gating network. The gating network is a lightweight convolutional network designed to compute adaptive weights for modality specific experts in a soft Mixture-of-Experts (MoE) framework. It processes fused multi-modal features using a two-layer convolutional block with batch normalization and Tanh activation to extract compact and discriminative representations. The design is flexible to support different modality combinations. The final output is produced by a 1×1 convolutional layer that maps the hidden features to a vector of length equal to the number of experts. This vector represents the raw gating logits, which can be converted into soft or hard weights to control the contribution of each expert during training and inference. Next, each expert takes it's input and gives the predictions. Finally, the output is generated by computing the weighted average of expert predictions using weights generated by the gating network. In soft Mixture-of-Experts (MoE) models, gating networks assign weights to different experts based on input relevance. However, without explicit regulation, the gating mechanism may collapse to favor a small subset of experts across most samples, under utilizing others. This imbalance limits the capacity of the model, reduces the diversity of learned representations, and can lead to overfitting of dominant branches. To mitigate this, a load balancing loss is introduced to encourage uniform expert activation across the batch. By penalizing skewed expert usage, this loss promotes better distribution of learning signals, ensures that all experts contribute meaningfully during training, and enhances generalization and robustness of the overall model. We employ a load balancing loss based on the coefficient of variation (CV), which measures the relative dispersion of expert usage across a mini-batch. Specifically, for each expert, the gating network produces a weight per sample; aggregating these weights over the batch yields an expert-wise usage distribution. The CV is computed as the ratio of the standard deviation to the mean of this distribution, capturing how unevenly the experts are selected. A high CV indicates that some experts dominate, while others are rarely used. CV-based loss penalizes this imbalance by minimizing CV, thereby encouraging more uniform expert activation and preventing mode collapse. This regularization improves learning dynamics, promotes specialization across experts, and leads to more effective and stable training in multi-modal settings. Equation~\ref{eq1} describes the CV based loss. Where $g_i$ represents the total gating weight assigned to expert $i$ in a batch, capturing how much that expert is used. The term $\mu = \frac{1}{N_E} \sum_{i=1}^{N_E} g_i$ is the mean gating load in all $N_E$ experts, and $\sigma$ is the standard deviation of these gating loads, measuring the variability in expert usage. The ratio $\frac{\sigma}{\mu}$ is the coefficient of variation (CoV), and squaring it yields a normalized measure of imbalance. The scalar $\lambda_{\text{dyn}}$ is a hyperparameter that controls the influence of this loss term during training. 

\begin{figure}[ht]
  \centering
  \includegraphics[width=0.8\textwidth]{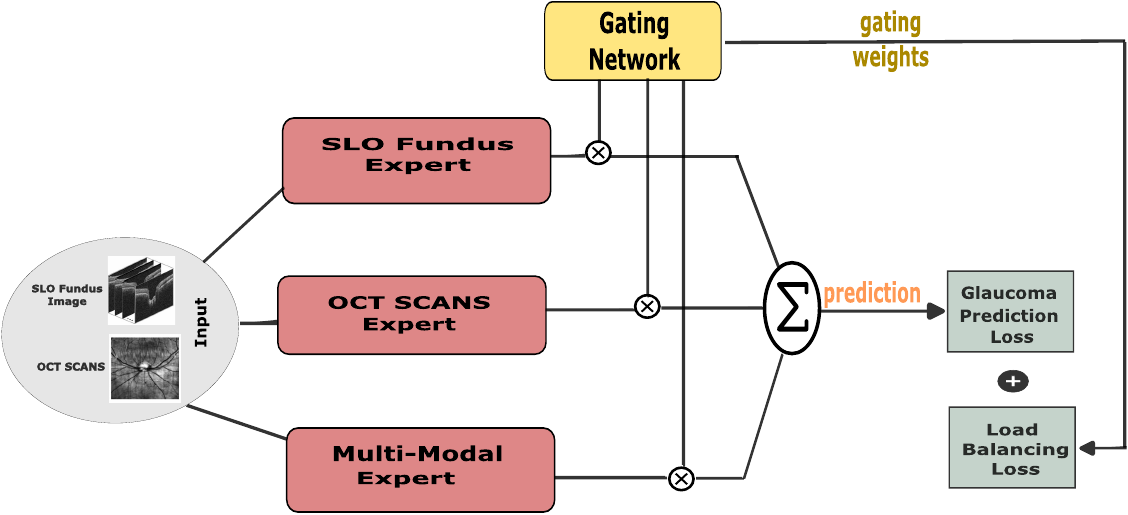}
  \caption{The overall architecture of SMoE. Here we take SLO Fundus and OCT Scans. Gating network generates the weights for the three experts. Each of the expert predictions is multiplied with respective weight and then added to get the final prediction.} 
  \label{fig4}
\end{figure}

\begin{equation}
\label{eq1}
\mathcal{L}_{\text{balance}} = \lambda_{\text{dyn}} \cdot \left( \frac{\sigma}{\mu} \right)^2 = \lambda_{\text{dyn}} \cdot \frac{1}{\mu^2 N_E} \sum_{i=1}^{N_E} (g_i - \mu)^2
\end{equation}

\subsection{Parameter Settings}

Following Fairvision settings [11], all convolutional neural network (CNN) models, including EfficientNet, DenseNet, and ResNet, were trained for 10 epochs with a batch size of 6 and an initial learning rate of $1 \times 10^{-4}$. For the Vision Transformer (ViT), we trained for 50 epochs using a batch size of 64, a base learning rate of $5 \times 10^{-4}$, a layer-wise learning rate decay of 0.55, and a drop path rate of 0.01. Hyperparameters for OPM, OGM, OPM+OGM, and CGGM were selected based on value ranges recommended in prior work. For our proposed SMoE method, we set the regularization coefficient for the load balancing loss to 0.001. All models were optimized using the AdamW optimizer with a weight decay of 0.01.

\section{Acknowledgements} 

The work was supported by NSF OIA-1946231, CNS-2117785, and OIA-2229752.

\section{Author contributions}
MS, SVC, and XH conceived the study. SVC and MS developed deep learning models. SVC and KR performed data processing, experiment, and analysis. MS, KR, and XH contributed materials and clinical expertise. MS  and XH supervised the work. All authors wrote and revised the manuscript. All authors have read and approved the manuscript.

\section{Competing Interests}

The authors declare no competing interests.



\end{document}